\documentclass[conference]{IEEEtran}
\IEEEoverridecommandlockouts
\usepackage{cite}
\usepackage{amsmath,amssymb,amsfonts}
\usepackage{graphicx}
\usepackage{textcomp}
\usepackage{xcolor}
\usepackage{tikz}
\usepackage{bm}
\usepackage{url} 
\usetikzlibrary{calc,arrows.meta, positioning, shapes.geometric}

\def\BibTeX{{\rm B\kern-.05em{\sc i\kern-.025em b}\kern-.08em
    T\kern-.1667em\lower.7ex\hbox{E}\kern-.125emX}}
\begin{document}

\title{GraphNNK - Graph Classification and Interpretability\thanks{This work was supported by the Montenegrin Ministry of Education, Science and Innovations under project DPG ``Data Processing on Graphs'', and by the Montenegrin Academy of Sciences and Arts. We acknowledge the EuroHPC Joint Undertaking for awarding this project access to the EuroHPC supercomputer LEONARDO, hosted by CINECA (Italy) and the LEONARDO consortium through an EuroHPC Development Access call.}

}
\author{
\IEEEauthorblockN{1\textsuperscript{st} {\v Z}eljko Bolevi\'{c}}
\IEEEauthorblockA{\textit{Faculty of Electrical Engineering} \\
\textit{University of Montenegro}\\
Podgorica, Montenegro \\
zeljkobolevic@ucg.ac.me}
\and
\IEEEauthorblockN{2\textsuperscript{nd} Milo\v{s} Brajovi\'{c}}
\IEEEauthorblockA{\textit{Faculty of Electrical Engineering} \\
\textit{University of Montenegro}\\
Podgorica, Montenegro \\
milosb@ucg.ac.me}
\\[1ex] 
\IEEEauthorblockN{4\textsuperscript{th} Ljubi\v{s}a Stankovi\'{c}}
\IEEEauthorblockA{\textit{Faculty of Electrical Engineering} \\
\textit{University of Montenegro}\\
Podgorica, Montenegro \\
ljubisa@ucg.ac.me}
\and
\IEEEauthorblockN{3\textsuperscript{rd} Isidora Stankovi\'{c}}
\IEEEauthorblockA{\textit{Faculty of Electrical Engineering} \\
\textit{University of Montenegro}\\
Podgorica, Montenegro \\
isidoras@ucg.ac.me}
}

\maketitle

\begin{abstract}

Graph Neural Networks (GNNs) have become a standard approach for learning from graph-structured data. However, their reliance on parametric classifiers (most often linear softmax layers) limits interpretability and sometimes hinders generalization. Recent work on interpolation-based methods, particularly Non-Negative Kernel regression (NNK), has demonstrated that predictions can be expressed as convex combinations of similar training examples in the embedding space, yielding both theoretical results and interpretable explanations.
\end{abstract}

\section{Introduction}
Graph-structured data appear naturally in numerous application domains, including molecular property prediction, citation networks, and social analysis \cite{stankovic2020part1, stankovic2020part2, stankovic2020part3}. To address these tasks, Graph Neural Networks (GNNs) have become powerful tools for representation learning \cite{b2, b3, stankovic202x_gcn_basis}. Among them, the Graph Isomorphism Network (GIN) is seen as one of the most expressive architectures for graph classification \cite{b4}.Despite their success, GNNs typically use parametric softmax classifiers, limiting interpretability. Although effective, such classifiers offer limited interpretability, which motivates the development of alternative approaches that provide more transparent decision mechanisms \cite{b5,b6}. 

Recent advances in explainability for GNNs include methods such as local subgraph identification \cite{b5}, taxonomic surveys of attribution techniques \cite{b6}, and hierarchical pooling strategies \cite{b7}. However, many of these approaches introduce additional complexity or may compromise predictive performance. In parallel, non-parametric, neighbor-based inference methods have been explored as a complementary direction. Classical nearest neighbor algorithms and their deep learning variants \cite{b9} demonstrate the potential of example-driven classification, while more recent formulations such as Non-Negative Kernel regression (NNK) \cite{b1} extend this idea by constructing geometrically consistent interpolations with theoretical guarantees on stability and generalization.

In this work, a GIN model is trained to learn graph embeddings, after which its parametric classification layer is replaced by the NNK interpolator applied in the learned embedding space. The goal is to examine whether a non-parametric inference scheme can serve as an effective alternative to softmax-based classifiers. Experiments on the NCI1\footnote{\url{https://chrsmrrs.github.io/datasets/docs/datasets/}} dataset show that NNK can achieve competitive or improved accuracy when applied to well-optimized embeddings, while its performance decreases when embedding quality deteriorates. These observations suggest that the effectiveness of NNK strongly depends on the geometric consistency of the learned representations and on the choice of the model used for inference.

The remainder of this paper is as follows. Section II outlines the theoretical foundations underlying GNNs, GIN, and the NNK regression framework. The proposed architecture is then described in Section III and evaluated through experiments in Section IV. The paper concludes with a discussion of results and potential directions for future work in Section V.

\section{Background Theory}

\subsection{Graph Neural Networks}
Relations between entities are commonly modeled using universal data structures known as graphs. 
A graph $\mathcal{G} = (\mathcal{V}, \mathcal{E})$ consists of a set of nodes $\mathcal{V}$ and edges $\mathcal{E}$, where edges describe pairwise relationships. 

GNNs learn node and graph representations by iteratively updating each node’s features based on its local neighborhood, similar to how CNNs update pixel features using information from nearby pixels.
This process, known as \emph{message passing} \cite{b10}, consists of two steps, 
an \emph{aggregation} of information from neighboring nodes and an \emph{update} of the node’s own state
\begin{equation}
\mathbf{h}_v^{(k)} = \text{update}\Big(\mathbf{h}_v^{(k-1)}, \text{aggregate}\{\mathbf{h}_u^{(k-1)} : u \in \mathcal{N}(v)\}\Big),
\end{equation}
where $\mathbf{h}_v^{(k)}$ is the embedding of node $v$ at the $k$-th layer, operator $\operatorname{aggregate(\cdot)}$ is a permutation-invariant function (e.g., \texttt{sum}, \texttt{mean}, \texttt{max}) that combines messages from neighbors $u \in \mathcal{N}(v)$, and operator $\operatorname{update(\cdot)}$ (e.g., an MLP) fuses the aggregated message with the previous state of node $v$. 
This allows information to propagate across $k$-hop neighborhoods as layers are stacked.

For a given node $v \in \mathcal{V}$, its neighborhood $\mathcal{N}(v)$ is defined as the set of all nodes directly connected to $v$ by an edge, 
\[
\mathcal{N}(v) = \{\, u \in \mathcal{V} \;|\; (u,v) \in \mathcal{E} \,\},
\]
that is, the immediate one-hop neighbors of $v$ in the graph.

\subsection{Graph Isomorphism Network (GIN)}
Among many GNN architectures, the GIN architecture in \cite{b4} is seen to be a highly expressive model for graph classification. Its update rule is defined as
\begin{equation}
\mathbf{h}_v^{(k)} = \operatorname{MLP}^{(k)}\!\Big((1+\epsilon)\, \mathbf{h}_v^{(k-1)} + \sum_{u \in \mathcal{N}(v)} \mathbf{h}_u^{(k-1)} \Big),
\vspace{-1mm}
\end{equation}
where $\epsilon \in \mathbb{R}$ is a learnable scalar parameter. 
Here, $\operatorname{MLP}^{(k)} : \mathbb{R}^{d_{k-1}} \to \mathbb{R}^{d_k}$ denotes a standard feed-forward neural network parameterized by a sequence of weight matrices $\mathbf{W}_i^{(k)} \in \mathbb{R}^{d_i \times d_{i-1}}$ and bias vectors $\mathbf{b}_i^{(k)} \in \mathbb{R}^{d_i}$, followed by nonlinear activations
\[
\operatorname{MLP}^{(k)}(x)\! = \!\sigma_{L}\big(\mathbf{W}_{L}^{(k)}\,\sigma_{L-1}(\dots\sigma_{1}(\mathbf{W}_{1}^{(k)}x + \mathbf{b}_{1}^{(k)})\dots) + \mathbf{b}_{L}^{(k)}\big).
\]
 Here, $\sigma_i(\cdot)$, $i=1,2,\dots,L$ denote activation functions and $L$ is the total number of layers.
The operator
$
\sum_{u \in \mathcal{N}(v)} \mathbf{h}_u^{(k-1)}
$
is referred to as \emph{sum aggregation} -- it collects the representations of all one-hop neighbors of node $v$ and combines them by element-wise summation. 
Sum aggregation is a permutation invariant function in the multiset $\{\mathbf{h}_u^{(k-1)} : u \in \mathcal{N}(v)\}$, ensuring that the result does not depend on any arbitrary ordering of neighbors. 
To obtain a graph-level representation, node embeddings from the final layer are combined using a pooling operator such as \emph{sum pooling} or \emph{mean pooling} \cite{b11}.

\subsection{Graph Classification}
The task of graph classification requires assigning a label to an entire graph. 
The standard pipeline consists of:
\begin{enumerate}
    \item training a GNN (e.g., GIN) to obtain node embeddings,
    \item pooling node embeddings to obtain a fixed-size graph embedding vector,
    \item passing the embedding through a classifier (commonly a softmax linear layer).
\end{enumerate}
In this work, instead of relying on a standard linear classifier, we employ interpolation-based classifiers, namely Non-Negative Kernel regression (NNK), to investigate accuracy and interpretability.

\subsection{Non-Negative Kernel Regression (NNK)}

The NNK regression is an interpolation-based, example-driven approach that expresses the prediction for a given sample as a non-negative linear combination of its neighboring samples in a reproducing kernel Hilbert space (RKHS). The main idea is that a data point should be represented only through its most informative neighbors, those that contribute new, linearly independent directions in the feature space, thus producing stable and interpretable local interpolations.

Let $\mathbf{x} \in \mathbb{R}^d$ denote a query sample, i.e., the point (a graph represented by its embedding vector) for which we want to estimate the output label, where $d$ represents the dimensionality of the graph embedding space. Let $\mathcal{S}=\{\mathbf{x}_1, \mathbf{x}_2, \dots, \mathbf{x}_k\}$ be its candidate neighborhood (for instance, the $k$-nearest training samples in the embedding space). Each sample $\mathbf{x}_i$ is mapped into a high-dimensional feature space via a kernel-induced function $\boldsymbol{\phi}(\cdot)$, and the corresponding matrix of neighbor embeddings is denoted as $\boldsymbol{\Phi}_\mathcal{S} = [\,\boldsymbol{\phi}(\mathbf{x}_1)\ \boldsymbol{\phi}(\mathbf{x}_2)\ \dots\ \boldsymbol{\phi}(\mathbf{x}_k)\,]$. In the context of graph classification, each input sample $\mathbf{x}_i$, $i=1,2,\dots, k$, corresponds to a graph-level embedding produced by the GNN network.

The goal of NNK is to reconstruct the feature representation of the query point $\boldsymbol{\phi}(\mathbf{x})$ as a linear combination of its neighbors
\begin{equation}
\boldsymbol{\phi}(\mathbf{x}) \approx \boldsymbol{\Phi}_\mathcal{S} \boldsymbol{\theta},
\end{equation}
where the coefficient vector $\boldsymbol{\theta} \in \mathbb{R}^k$ contains non-negative reconstruction weights. These weights quantify how much each neighbor contributes to explaining (or interpolating) the query point in the feature space. The optimal weights are obtained by solving the constrained least-squares problem, i.e.
\begin{equation}
\min_{\boldsymbol{\theta} \ge 0} \; \|\boldsymbol{\phi}(\mathbf{x}) - \boldsymbol{\Phi}_\mathcal{S}\boldsymbol{\theta}\|^2 .
\label{eq:nnk_obj}
\end{equation}
By applying the kernel method in \cite{b1}, the reconstruction error can be expressed in terms of kernel evaluations, yielding the quadratic form
\begin{equation}
\min_{\boldsymbol{\theta} \ge 0}\; 1 - 2\,\boldsymbol{\theta}^\top \mathbf{K}_{\mathcal{S},*} + \boldsymbol{\theta}^\top \mathbf{K}_{\mathcal{S},\mathcal{S}}\,\boldsymbol{\theta},
\end{equation}
where $\mathbf{K}_{\mathcal{S},*} = [\,K(\mathbf{x}_1,\mathbf{x}), \dots, K(\mathbf{x}_k,\mathbf{x})\,]^\top$ is the vector of kernel similarities between the query point and its neighbors, and $\mathbf{K}_{\mathcal{S},\mathcal{S}}$ is the $k \times k$ kernel matrix between all pairs of neighbors in $\mathcal{S}$.

After optimization, only a subset of coefficients in $\boldsymbol{\theta}$ are typically nonzero, indicating which neighbors are most relevant for locally describing $\mathbf{x}$. Let $\hat{k}$ denote the number of such active neighbors. For regression, the predicted value is given by the convex combination
\begin{equation}
\hat{\eta}(\mathbf{x}) = \sum_{i=1}^{\hat{k}} w_i\, y_i,
\qquad
w_i = \frac{\theta_i}{\sum_{j=1}^{\hat{k}} \theta_j},
\label{eq:nnk_unbiased}
\end{equation}
where $y_i$ are the known labels of the selected training samples, and $w_i$ are normalized non-negative weights satisfying $\sum_i w_i = 1$. For classification tasks, these weights represent the contribution of each active neighbor to the decision, and class probabilities are obtained by convexly interpolating the one-hot label vectors of the neighbors. 

Intuitively, the NNK weights $\boldsymbol{\theta}$ define how the query point is expressed as a mixture of its most informative neighbors. The non-negativity constraint ensures that the interpolation remains within the local convex region of the data manifold, which in turn leads to both interpretability and better generalization. 

The geometry of NNK provides additional insight: only those neighbors that contribute new, linearly independent directions remain active with nonzero weights, forming a local convex polytope around the query point. A necessary and sufficient condition for two candidates \(\bm{x_j}\) and \(\bm{x_k}\) to simultaneously appear as NNK neighbors of \(\bm{x}\) can be expressed through the kernel ratio interval (KRI),
\begin{equation}
K_{j,k} \; < \; \frac{K_{i,j}}{K_{i,k}} \; < \; \frac{1}{K_{j,k}},
\label{eq:kri}
\end{equation}
which characterizes the geometric relationships among the neighbors and identifies the most influential training samples \cite{b1}. As a result, each prediction can be directly interpreted through the specific examples that actively contribute to it.

The optimization yields a sparse weight vector \(\boldsymbol{\theta}\), so only a few neighbors get nonzero weights. This sparsity lets NNK reconstruct the query using the most informative neighbors, forming a compact convex polytope in the kernel feature space, where smaller, localized regions lead to smoother and more stable predictions.

\section{Proposed Architecture}

The proposed framework combines the graph neural networks, specifically the GIN, with the interpretability of NNK as the final classification layer.

\subsection{Overview}
The architecture is divided into two stages:
\begin{enumerate}
    \item \emph{Representation learning with GIN.} A GIN model is trained on the training set to learn graph embeddings that capture both structural and attribute information. After training, the GIN acts as a feature extractor that maps each graph into a low-dimensional vector representation.
    \item \emph{Classification with NNK.} Instead of relying on a parametric softmax layer, classification is performed using NNK interpolation applied directly on the learned graph embeddings. 
\end{enumerate}

\begin{figure}[htbp]
\centering
\includegraphics[width=0.5\textwidth]{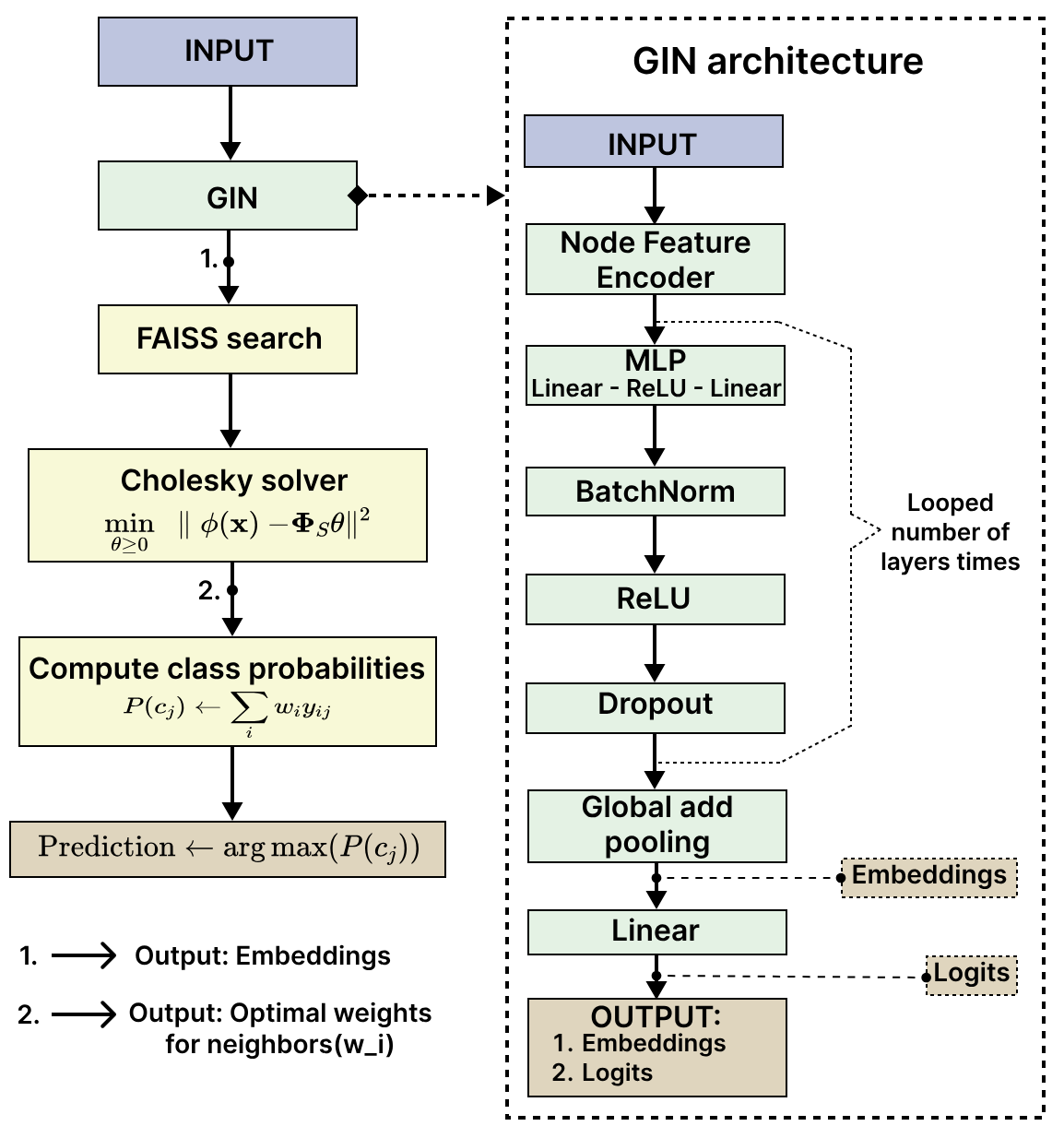}
\caption{Proposed architecture for graph classification.}
\label{fig:architecture}
\end{figure}

As illustrated in Fig.~\ref{fig:architecture}, the proposed architecture builds upon the standard GIN framework, which is trained with the primary goal of extracting meaningful hidden representations from the training graphs. 

After the training phase, the GIN architecture is extended with the NNK classifier. During evaluation, the NNK classifier operates in the embedding space: for each test graph, it first uses FAISS \cite{b14} to retrieve its nearest neighbors among the training graphs. Here, the term \emph{nearest neighbors} refers to the training graphs whose embeddings are most similar to the embedding of the test graph, according to the distance metric defined by FAISS (e.g., Euclidean or cosine similarity).

Once the relevant neighbors are identified, a Cholesky-based solver is applied to solve the quadratic optimization problem defined in (4). This procedure yields the coefficients $w_i$, which represent the relative contribution of each selected neighbor to the interpolation. Based on these coefficients, class probabilities are computed by convexly combining the labels of the active neighbors, and the test graph is assigned to the class with the highest resulting probability.

\subsection{Interpretability}
An important advantage of this design is its interpretability.
Unlike standard softmax classifiers, NNK explicitly identifies which training examples contribute to the prediction, along with their interpolation weights. This provides a transparent, example-based explanation of each decision, where only a subset of neighbors with non-zero weights influences the final outcome.
A key benefit of this approach is that interpretability can be achieved without compromising accuracy or efficiency, unlike many existing methods that offer interpretability at the cost of performance \cite{b13,b9}.

\section{Experimental Results}

\begin{figure}[t]
    \centering
    \includegraphics[width=\columnwidth]{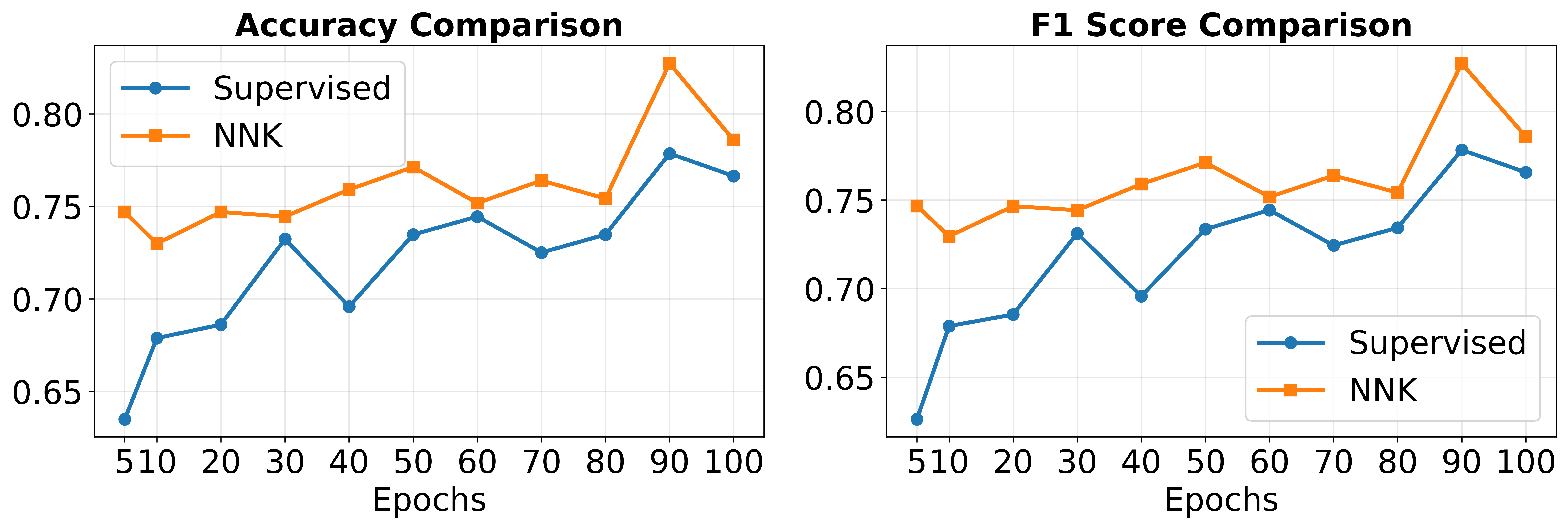}\\[2mm]
    \includegraphics[width=\columnwidth]{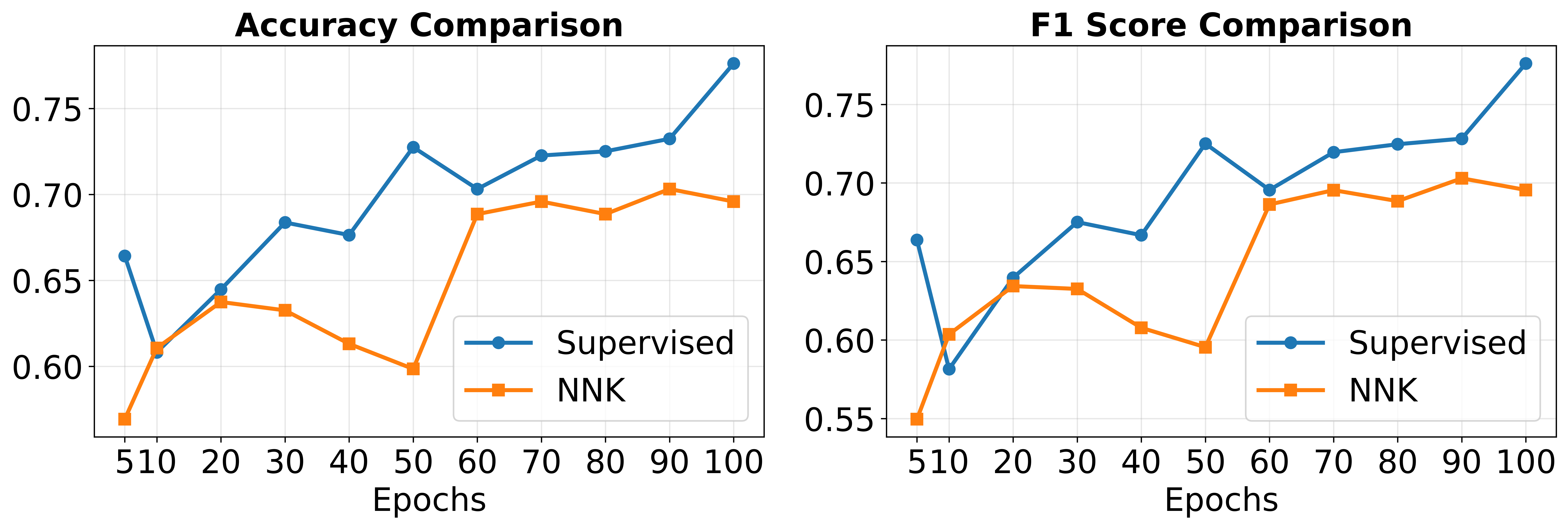}
    \caption{Supervised vs. NNK comparison: best validation checkpoint (top) and last training snapshot (bottom). Each pair shows accuracy (left) and F1 score (right).}
    \label{fig:comparison}
\end{figure}

All experiments were performed on the NCI1 dataset from the TU collection, using GIN as the base architecture.  Table~\ref{tab:hyperparameters} summarizes the main hyperparameters used during training. 

\renewcommand{\arraystretch}{1.2}
\begin{table}[b]
\caption{Main hyperparameters used in the experiments.}
\centering
\begin{tabular*}{\columnwidth}{@{\hspace{2mm}}l@{\extracolsep{\fill}}c@{\hspace{2mm}}}
\hline\hline
Parameter & Value \\
\hline
Number of layers ($L$)        & 5 \\
Hidden dimension ($d_h$)      & 128 \\
Dropout probability ($p_d$)   & 0.5 \\
Learning rate ($\eta$)        & $1\times10^{-3}$ \\
Batch size ($B$)              & 128 \\
Number of neighbors ($k_{\mathrm{FAISS}}$) & 50 \\
Sparsity threshold ($\tau_{\mathrm{edge}}$) & $1\times10^{-10}$ \\
\hline\hline
\end{tabular*}
\label{tab:hyperparameters}
\end{table}

Since the NCI1 dataset does not provide explicit node features, each node was assigned a feature corresponding to its degree, following the common practice for datasets without intrinsic node attributes.

Two approaches were compared under identical training conditions.  
The first is a \emph{supervised baseline}, where a linear classification layer is trained on top of the GIN graph embeddings using cross-entropy loss.  
The second is the \emph{non-parametric NNK classifier}, applied on the learned embeddings after the GIN training phase. 

Fig. \ref{fig:comparison}  presents the evaluation results for two different model checkpoints: the best validation checkpoint (top) and the final training snapshot (bottom), respectively.
When evaluated at the best validation checkpoint, NNK consistently outperforms the supervised approach across all epochs. For instance, at epoch 90, accuracy increases from 0.7786 (supervised) to 0.8273 (NNK). Macro-F1 follows the same trend, indicating that the performance gains are not limited to dominant classes. In contrast, when the evaluation is performed on the last training snapshot, NNK generally underperforms relative to the supervised classifier, confirming its sensitivity to the quality and stability of the embedding space. 

\section{Conclusion}
This paper explored the integration of NNK with GIN for graph classification. By replacing the standard parametric softmax layer with a non-parametric NNK interpolator, the proposed approach enables predictions to be expressed as convex combinations of informative training samples, thus enhancing interpretability without requiring additional parameter training.  
Experimental results on the NCI1 dataset showed that NNK consistently outperforms the supervised baseline when applied to well-optimized graph embeddings, highlighting its ability to exploit the local geometric structure of the representation space.   
Future work will explore adaptive non-parametric classifiers.

\end{document}